\documentclass[11pt]{article}

\usepackage[a4paper,margin=2.5cm]{geometry}
\usepackage{graphicx}
\usepackage{multirow}
\usepackage{amsmath,amssymb,amsfonts}
\usepackage{amsthm}
\usepackage{mathrsfs}
\usepackage{xcolor}
\usepackage{textcomp}
\usepackage{booktabs}
\usepackage{placeins}
\usepackage{array}
\usepackage{longtable}
\usepackage{enumitem}
\usepackage[font=small,labelfont=bf]{caption}
\usepackage[numbers,sort&compress]{natbib}
\usepackage[colorlinks=true,allcolors=blue!50!black]{hyperref}

\graphicspath{{figures/}}

\newcommand{\sinote}[1]{Supplementary~Note~\ref{#1}}

\newcommand{\mainfig}[1]{Fig.~\ref{#1}}

\title{Earth embeddings are decomposable records of the land surface}
\author{\begin{tabular}[t]{c}
Robin Young\thanks{Corresponding author: ray25@cam.ac.uk}\qquad Srinivasan Keshav\thanks{sk818@cam.ac.uk}
\\[0.6em]
\small Department of Computer Science and Technology, University of Cambridge,\\
\small William Gates Building, Cambridge CB3 0FD, UK
\end{tabular}}
\date{}

\begin{document}
\maketitle

\begin{abstract}
Ecological monitoring, carbon accounting and biodiversity assessment increasingly rely on annual Earth embeddings---learned vectors that compress a year of satellite observation for every ten-meter pixel on Earth, consumed in place of the imagery itself. What these vectors mean has not been established. Here we decompose two widely used embedding products and show that they contain individual ecological features, including fire, forest clearing and crop type, validated against independent reference data on events across six continents. The recovered concepts date abrupt canopy loss to the first post-disturbance growing season and resolve severity well enough to separate prescribed fire from wildfire. Small fires that recover quickly are detectable in one product but not the other, which instead separates crops more strongly.
\end{abstract}


\section{Introduction}
\label{sec:intro}

The land surface of Earth is now observed, compressed and distributed as learned vectors \cite{klemmer2025earthembeddings}. Geospatial foundation models (GeoFMs) \cite{prithvi2023,presto2023,satmae2022,dofa2024,galileo2025,herzog2026OlmoEarth} ingest a year of multi-sensor satellite imagery for every ten-meter pixel and produce a single annual embedding. Two such models \cite{tessera2025, alphaearth2025} now provide these embeddings as downloadable precomputed global data products intended to replace raw imagery as the standard input to downstream analysis. Embedding products are already used as input features for land-cover mapping, agricultural monitoring, ecological assessment and carbon accounting \cite{klemmer2025earthembeddings,ma2026alphaearth,joseph2026biome, lisaius2026croptype, ball2026geospatial, young2026belowground, ryan2026AlphaEarthWetland, karimi2026Miankaleh}.

Good downstream accuracy tells us that a task’s target can be recovered by a model trained on the embeddings, but not what the embedding itself represents, or which part of it does. A scientist choosing an embedding product for modeling has no way to know whether the product contains the phenomenon they are interested in. Klemmer et al. name interpretability as one of three open research gaps for Earth embeddings. ``We don't know which part of an Earth embedding represents a given ground condition, such as a tree or a building--or if the embedding captures these ground conditions at all'' \cite{klemmer2025earthembeddings}.

Existing representation-level studies correlate embedding dimensions with environmental covariates \cite{rahman2026physically, she2026pico}, probe the linear accessibility of reanalysis variables \cite{mocharla2026probing}, characterize representational geometry \cite{romero2026layerwise}, or decompose location encoders \cite{betti2026whatsinanearthembedding, cole2023sinr}. These tell us what an embedding space looks like or what a location encoder knows about geography, but location encoders are conditioned on coordinates, so anything recovered from them is a fact about where a place is, and not about what happened there. Model evaluation otherwise typically runs through downstream task accuracy \cite{geobench2023, pangaea2024}, a practice the benchmark authors themselves have questioned \cite{geobench2_2025}. Explainability work in remote sensing has generally focused on post-hoc attribution for supervised classifiers \cite{hohl2024xai} rather than on the contents of a learned representation. 

What is missing is a way to isolate components of an embedding that correspond to specific surface events, and to verify them against independent records. This raises several questions. Do Earth embeddings contain features that correspond to identifiable surface phenomena and can be verified against independent maps? If so, what do those features record about a disturbance, such as when it occurred, how severe it was or what caused it? And do different embedding products, trained on largely the same observations of Earth, contain the same phenomena?

Here we decompose the two publicly available global embedding products, Tessera \cite{tessera2025} and AlphaEarth \cite{alphaearth2025}, with sparse autoencoders (SAEs) \cite{cunningham2023sparse,bricken2023monosemanticity}, a method developed to interpret language models. An SAE rewrites each embedding as a weighted combination of a few features, chosen from a learned set with many more features than the embedding has dimensions. Because each pixel can use only a few features, each feature tends to specialize on one surface condition, unlike principal components, which mix many. We test whether a feature records a phenomenon by when it appears, since a feature that represents fire should activate on burned pixels in the year of a mapped fire or after, and not before.

We applied this test to 3{,}940 mapped fires, forest disturbances, clearings, landslides, newly formed water bodies and crop fields on six continents, using public reference products \cite{picotte2020MTBS, eidenshink2007MTBS, GA2024bushfire, vianasoto2025EFDA, vianasoto2024EFDAdata, forzieri2020FORWIND, hansen2013, reiche2021RADD, boryan2011CDL, berti2025RER2023, ma2024RLBJ, fu2025RLZX, pekel2016GSW}. We focus on disturbances and crops because they change at known times and are independently mapped at continental scale, which lets each feature be tested against when an event occurred. Features selected on a subset of events dated 85\% of the remaining disturbances to their expected time window, above random feature sets of the same size in every region. The features resolve severity well enough to separate prescribed fire from wildfire, and they place each disturbance in the calendar year containing its first post-disturbance growing season. Tessera and AlphaEarth, trained on largely the same Sentinel observations, retain different phenomena. Small fires that recover quickly are detectable in Tessera's embeddings but not in AlphaEarth's, whereas corn and soy are better separated in AlphaEarth's, possibly because their training objectives treat time differently.

\section{Results}
\label{sec:results}

We first demonstrate the method on one fire and one cropland tile, then test it on 3{,}940 mapped events across 16 settings on six continents (Section~\ref{sec:scale}).

\subsection{Earth embeddings contain ecological features}
\label{sec:decompose}

\begin{figure}[htbp]
\centering
\includegraphics[width=\textwidth,keepaspectratio]{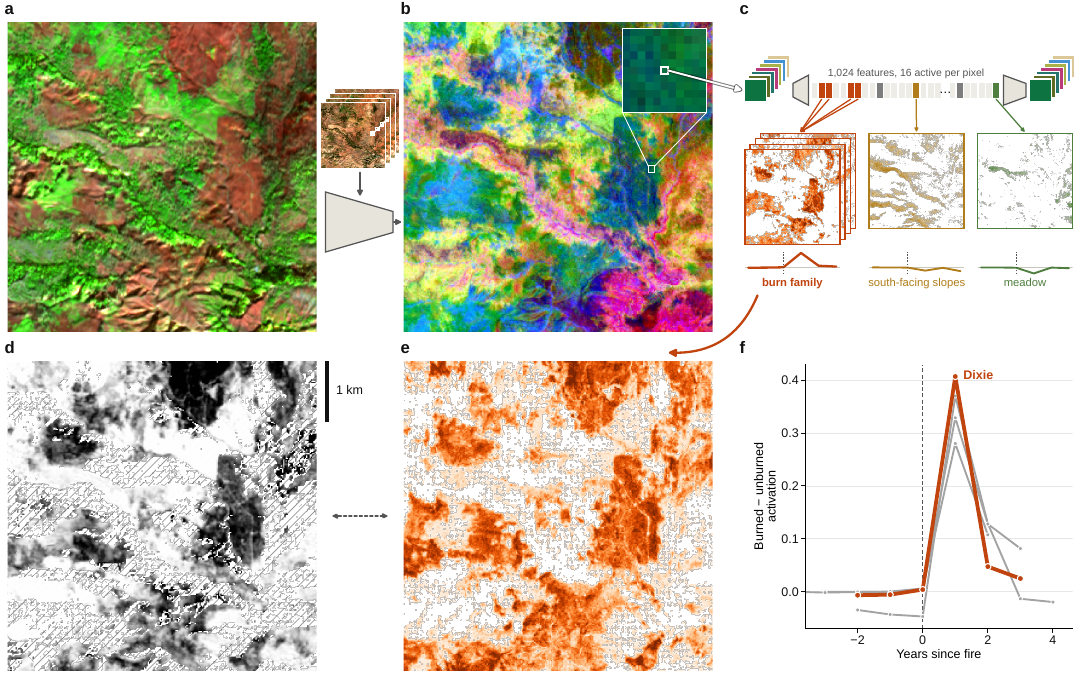}
\caption{\textbf{Sparse decomposition recovers a fire from Earth embeddings.} \textbf{a}, The Dixie Fire (California) in Sentinel-2 shortwave infrared imagery after containment. \textbf{b}, A year of Sentinel-1 and Sentinel-2 observations of the same area, compressed into Tessera's 128-dimensional annual embedding (first three principal components. Inset individual 10m pixels). \textbf{c}, A sparse autoencoder decomposes each pixel's embedding. Below we show a few features' activation maps. The burn features family (32 features) is the set selected from burned-minus-unburned contrast after the fire. Other illustrative features are south-facing slopes (91\% of its active pixels face south, from the Copernicus DEM, against 55\% of the tile) and a meadow feature whose activation rises from 2\% of forest pixels more than 300m from a meadow to 51\% of pixels more than 100m inside meadows (ESA WorldCover). \textbf{d}, Burn severity from the dNBR spectral index. \textbf{e}, Mean activation of the burn family, which separates burned from unburned pixels with AUC 0.914. 96.0\% of pixels with above-median activation are burned (dNBR $> 0.27$, pixels mapped as moderately or severely burned). \textbf{f}, Burned-minus-unburned family activation by year since fire, for the Dixie Fire (coral) and for three Sierra Nevada fires on which the family was not screened (gray lines are North Complex 2020, Caldor 2021, Mosquito 2022).}
\label{fig:decompose}
\end{figure}

To test whether embeddings contain verifiable features, we first decompose the embeddings over a single fire. The 2021 Dixie Fire burned across much of the northern Sierra Nevada (Fig.~\ref{fig:decompose}a). Fig.~\ref{fig:decompose}b shows the embeddings for the same area, with its first three principal components displayed as red, green and blue. The burn scar is visible in blue-green, so the embedding contains information about the fire, but that information is spread across many of the 128 dimensions, and no single dimension corresponds to the fire.

To find directions that do, we decompose the embeddings with a sparse autoencoder (SAE) \cite{gao2024scaling}. An SAE learns a dictionary of directions in embedding space, here $m = 1{,}024$ directions, eight times the number of embedding dimensions. It then rewrites each pixel’s annual embedding as a weighted sum of a few of these directions, with non-negative weights. We call each direction a feature and its weight at a pixel the feature’s activation. Because each pixel can use only a small number of features (a TopK SAE \cite{gao2024scaling}, Methods), features tend to specialize on distinct surface conditions instead of sharing variance as principal components do. Each feature therefore has its own activation map on the same area (Fig.~\ref{fig:decompose}c). We train the dictionary on embeddings alone, independent of the downstream task.

The decomposition produces 1,024 features, but it does not say which of them, if any, describe fire. To find them, we screen features by when their burn contrast appears. For each feature and each year, we compute its mean activation over burned pixels (dNBR $> 0.27$, Fig.~\ref{fig:decompose}d) minus its mean activation over unburned pixels in the same tile, which we call the \textit{contrast}. A feature that represents fire should show this contrast only after the fire. We select a feature if its contrast is large in the expected window (for western US fire, the year after the fire), small in every year before the fire, and has largely decayed by the third year after it (Methods). We call the features that pass this screen a feature family. On the Dixie tile, 32 of the 1,024 features pass (Fig.~\ref{fig:decompose}c).

Averaging the 32 features' activations gives one value per pixel, which reproduces the burn scar (Fig.~\ref{fig:decompose}e). Ranking pixels by family activation separates burned from unburned pixels with an AUC of 0.914, and 96.0\% of pixels with above-median activation are burned according to dNBR. The family was selected on a single tile, but if we apply it without modification to three other fires (North Complex 2020, Caldor 2021 and Mosquito 2022), its contrast also peaks in the first year after each fire (Fig.~\ref{fig:decompose}f). The embedding therefore appears to represent fire through a small set of identifiable features that behave consistently across the three fires tested. In the rest of the paper, we mark a held-out event as correctly dated when its family contrast is highest within its expected window, a year or pair of years fixed for each setting before screening (Methods).

\subsection{Features carry single meanings}
\label{sec:sparsity}

Why decompose embeddings at all, when a classifier trained on the raw embeddings could also have mapped the burn scar in Fig.~\ref{fig:decompose}? With a decomposition, each of its directions can be interpreted on its own. A raw embedding dimension can be informative without being interpretable, because its high and low values may correspond to unrelated surface types, as the next example shows.

\begin{figure}[htbp]
\centering
\includegraphics[width=\textwidth,keepaspectratio]{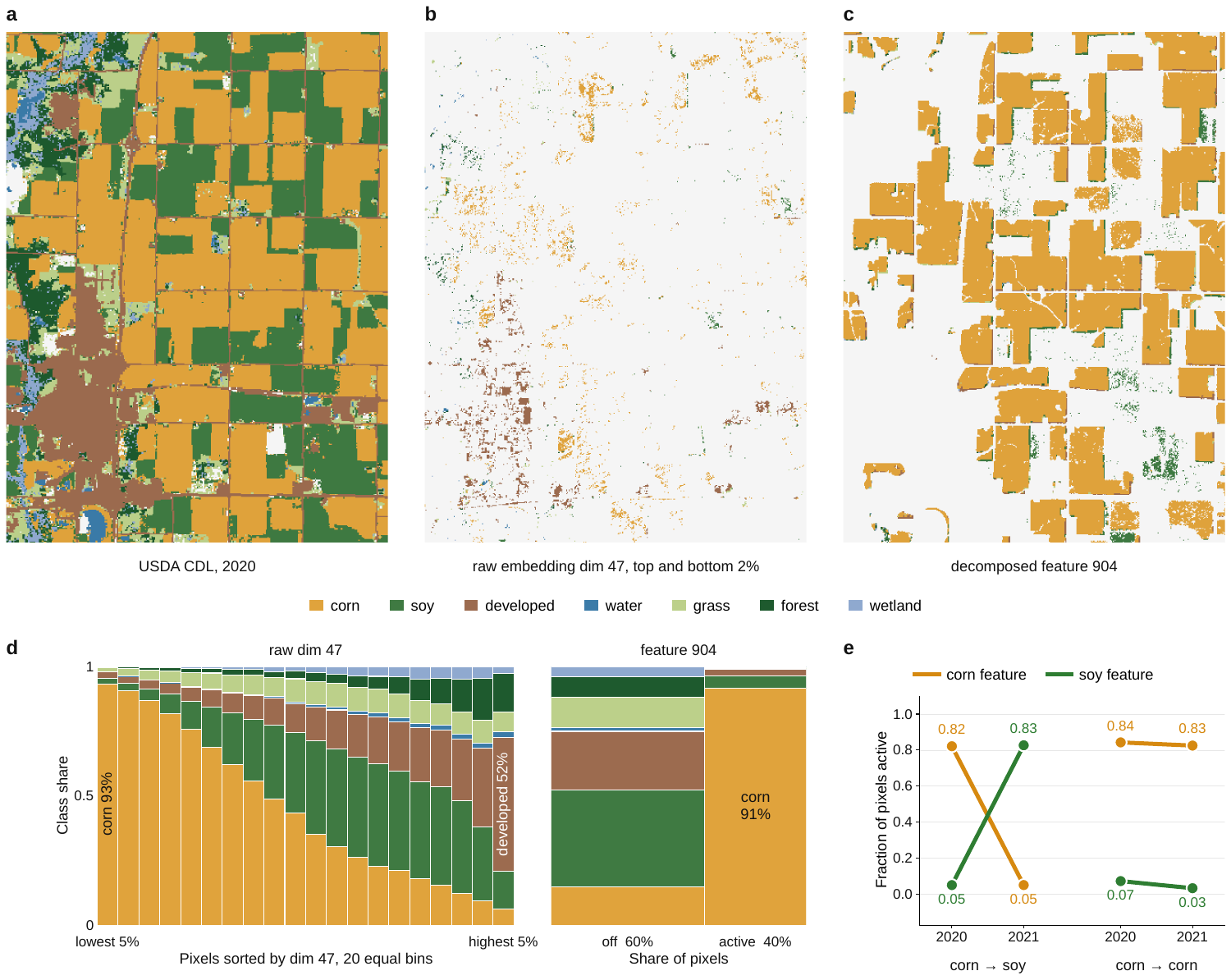}
\caption{\textbf{A decomposed feature identifies one land cover class.} \textbf{a}, USDA Cropland Data Layer (CDL) classes on one Iowa tile in 2020. \textbf{b}, Pixels in the top and bottom 2\% of raw embedding dimension 47, colored by CDL class. The bottom tail is mostly corn (93\%) and the top tail mostly developed land (68\%). \textbf{c}, Pixels where decomposed feature 904 is active, colored by CDL class. 91\% are corn, against 45\% across the tile overall. \textbf{d}, Class composition of the tile's pixels, ranked by their value on dimension 47 and split into 20 bins of equal pixel count (left), and of pixels where feature 904 (corn) is off or active (right). The dimension orders classes from corn through soy to developed land and identifies a single class only near its ends. The feature is active on 40\% of the tile, and 91\% of those pixels are corn. \textbf{e}, Share of pixels where the corn and soy features are active in 2020 and 2021, on pixels the CDL records as corn then soy (left, 66{,}881 pixels) and as corn in both years (right, 33{,}152 pixels).}
\label{fig:sparsity}
\end{figure}

Fig.~\ref{fig:sparsity} shows an example on a Corn Belt tile in Iowa, compared to the USDA Cropland Data Layer (CDL) \cite{boryan2011CDL}. Raw dimension 47 separates crops well. Among the 2\% of pixels with the highest values on this dimension, however, 68\% are developed land, while among the 2\% with the lowest values, 93\% are corn (Fig.~\ref{fig:sparsity}b,d). To interpret the dimension, we must know which end of the dimension a pixel lies on, and the two ends correspond to different dominant classes.

SAE features avoid this problem because any individual feature's activation is non-negative and zero on most pixels. Non-negativity alone does not guarantee that a feature corresponds to one class, a fact that we verify using the CDL. We selected the feature with the highest F1 score for CDL corn in 2020. Of the pixels where it is active, 91\% are corn, compared with 45\% of the tile overall, and no other class makes up a substantial share (Fig.~\ref{fig:sparsity}c). The corn feature identifies corn similarly well in 2020 and 2021 (F1 of 0.856 and 0.857), suggesting that it does not depend on year-specific conditions. The same feature is the best corn feature on 99 of 100 Corn Belt tiles (\sinote{note:dict}).

Corn and soy are commonly grown in rotation in the midwest US, so a field often alternates between them. On pixels the CDL records as switching from corn to soy between 2020 and 2021, the share of pixels where the corn feature is active falls from 82\% to 5\%, and the share for the soy feature rises from 5\% to 83\%. On pixels recorded as corn in both years, both shares stay nearly constant (Fig.~\ref{fig:sparsity}e). Across the corn belt, if each pixel is labeled by whichever feature is strongest, the label changes between years on 95.6\% of pixels the CDL records as rotating between corn and soy. The features follow the crop in the field each year instead of a fixed property of the place. The same corn and soy features are the best single features for their crops on 99 and 96 of 100 Corn Belt tiles, and outperform a fixed raw dimension on 71 and 98 of them (\sinote{note:dict}).

\subsection{Feature families date disturbances on six continents}
\label{sec:scale}

\begin{figure}[htbp]
\centering
\includegraphics[width=\textwidth,keepaspectratio]{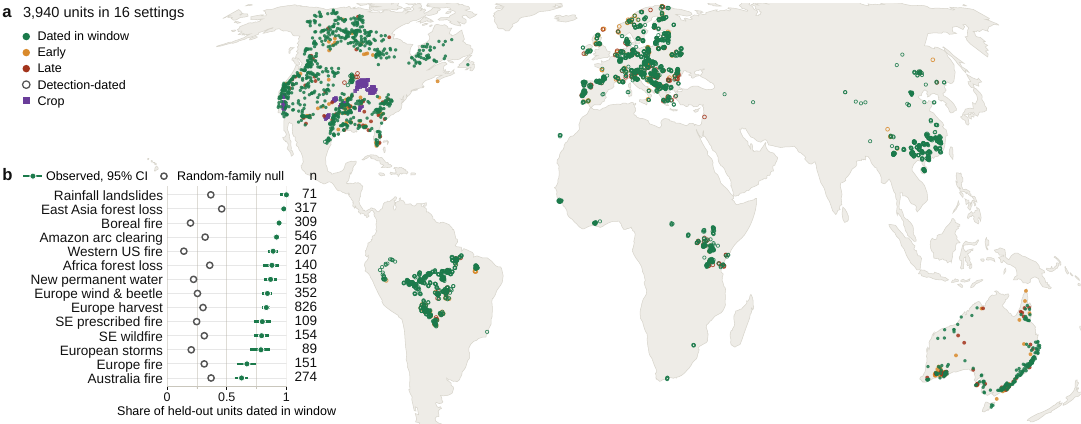}
\caption{\textbf{Validating decomposed feature families across 3{,}940 units in 16 settings on six continents.} \textbf{a}, Every unit, colored by its outcome of peaking in its setting's expected window, early or late. Filled markers are event-dated references and open markers detection-dated, which record different physical dates. Purple squares are crop units, which are classified. \textbf{b}, Held-out validation rate per dated setting (filled) with its 95\% confidence interval, against the median of a random-family null computed for that setting (open), with the number of held-out units ($n$). Rates and nulls are computed on the same held-out units. All 14 dated settings are above their nulls at $P \le 0.05$.}
\label{fig:scale}
\end{figure}

To test whether these features generalize beyond the examples above, we assembled 3{,}940 mapped events and tile-years (pixels of one disturbance type disturbed in one year within one tile), which we call units, spanning fires, harvest, windthrow, bark beetle damage, deforestation, landslides, newly formed water bodies and crops across six continents (Fig.~\ref{fig:scale}a). We organized these into 16 settings, each covering a disturbance type or crop system in one region and dated by a public reference product, such as MTBS fire perimeters in the western US, the European Forest Disturbance Atlas across 36 countries, Hansen forest loss maps in the Amazon, Africa and East Asia, or landslide inventories mapped after extreme rainfall in Italy and China (full list in Methods).

Purpose-built change detection algorithms already date disturbances well, and several of them produced the reference data we test against \cite{kennedy2010landtrendr,zhu2014ccdc,hansen2013}. Our question is what a general-purpose annual embedding encodes on its own, without any change-detection modeling. In each setting, we screened a feature family on a subset of units (typically its earliest year, see Methods) and tested it on all remaining units.

Because the dictionary contains 1,024 features, some may peak in the expected window by chance. We therefore compare each family against random sets with the same number of features, drawn from the same dictionary and scored on the same held-out units and windows (Methods). In every setting, the family exceeded its null ($P \le 0.05$, Fig.~\ref{fig:scale}b). Across the corpus, families dated 85\% of held-out disturbance events to their expected window, with per-setting rates ranging from 0.624 for Australian fire (null 0.369) to 1.000 for rainfall landslides (null 0.366). For the European Forest Disturbance Atlas, the family dated 83.2\% of 826 held-out harvest tile-years, against a null of 0.301. In the crop settings, which are classified rather than dated, 84 of 102 crop atlas tiles and 95 of 100 crop rotation tiles reach their held-out accuracy bar (Methods). 

A family screened on one year therefore dates disturbances in later years across every region and disturbance type we tested, which random feature sets of the same size do not. Rates are lowest where events span the calendar boundary or are small relative to the 10m pixel (Australian and European fire), and highest where disturbances are large and abrupt (landslides, forest loss in southern China, boreal fire). The families are not specific to vegetation loss as they also date land that becomes permanent water, mostly reservoirs filling and lakes rising (86.7\% of 158 held-out tiles, null 0.222). The embedding also holds families for surfaces that are not vegetation at all, such as solar farms and mining land, which we score by classification rather than dating (\sinote{note:detection}).

Validation rates remain above their nulls across a 16-fold range of the screening threshold and across dictionary seeds (\sinote{note:screen}). Families also transfer between continents. A winter wheat family learned in the US Plains identifies wheat in Europe as well as it does at home (AUC 0.80 in both), even though in raw embedding space European wheat lies closer to European cropland than to US wheat (\sinote{note:specificity}). A family screened on one year therefore dates disturbances in later years across every region and disturbance type we tested, which random feature sets of the same size do not.

\subsection{Feature timing follows phenology and vegetation recovery}
\label{sec:clock}

We next ask what the features record about a disturbance’s timing. The year in which a family's contrast peaks shows when a disturbance is most distinct in the annual record. Across the settings we analyzed, the peak activation happens in the calendar year that contains the first growing season after the disturbance. This is governed by two factors, namely when the event occurs relative to the local growing season, and how quickly vegetation responds afterward.

\begin{figure}[htbp]
\centering
\includegraphics[width=\textwidth,keepaspectratio]{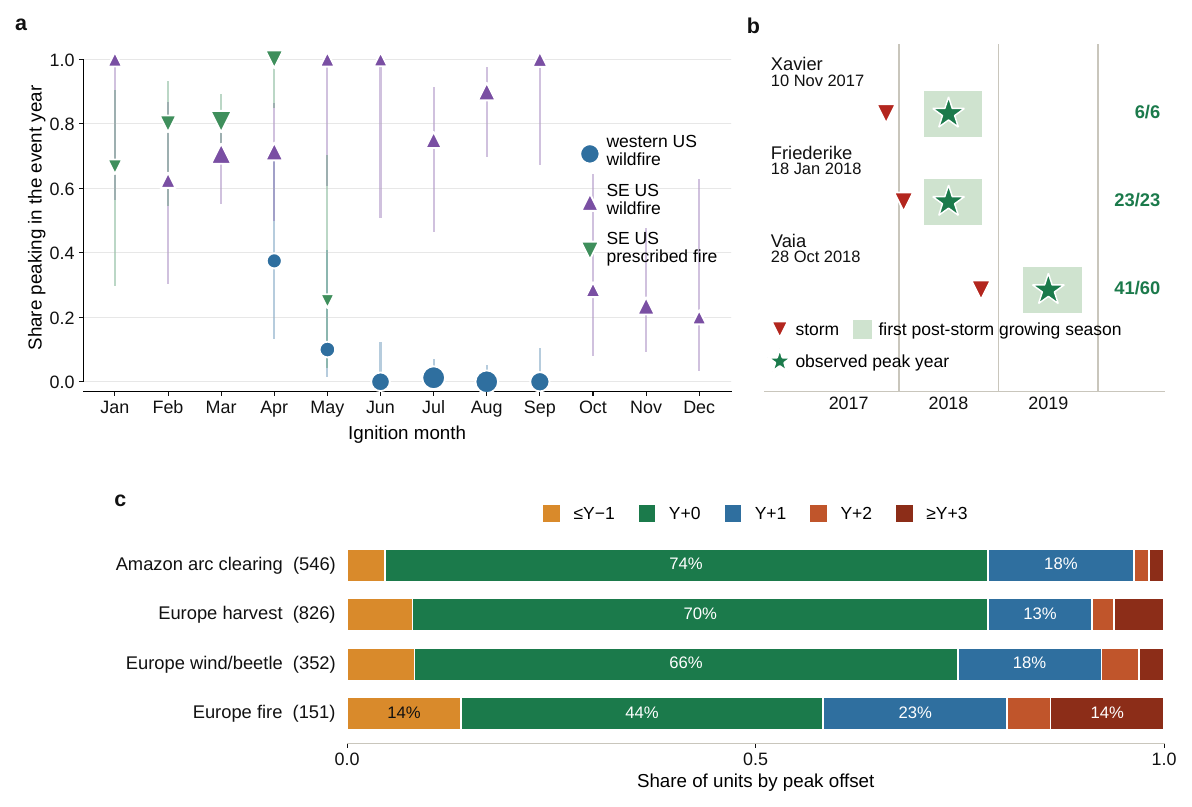}
\caption{\textbf{Disturbance features peak in the first growing season that shows the disturbed surface.} \textbf{a}, Share of units peaking in the event year, by ignition month, one curve per setting. Wilson 95\% intervals are shown, point area scales with $n$. Note each setting's growing season (western US April--October, SE US March--October). Western US wildfire peaks in the following year at nearly every ignition month, while SE US wildfire and prescribed fire mostly peak in the event year. Months with fewer than 4 events are dropped. \textbf{b}, Three windstorms in the same forest type with Xavier and Friederike ten weeks apart across a calendar boundary, and Vaia after the 2018 growing season, with markers at event dates, green bands marking each storm's first post-storm growing season, and stars the observed peak years. Friederike (23 of 23 tiles) and Xavier (6 of 6) peak in the predicted year, and Vaia in 41 of 60. \textbf{c}, Detection-dated settings, where the reference dates the first annual composite showing the disturbance.}
\label{fig:clock}
\end{figure}

Fire in the western and southeastern US shows the vegetation response factor. Only 2\% of western US wildfires peak in the event year, compared with 80\% of southeastern US prescribed fires (Fig.~\ref{fig:clock}a). Western montane conifer forests do not regrow in the remainder of the season in which they burn \cite{pausas2014evolutionary}, so their first post-fire growing season is the following year. Compare this to southeastern understory, which regrows within weeks \cite{glitzenstein1995effects}, so the event year already contains it. Southeastern prescribed fires peak in the event year regardless of ignition month, consistent with rapid regrowth after any ignition date.

Xavier (10 November 2017), Friederike (18 January 2018) and Vaia (late October 2018) occurred in similar conifer forests in central Europe and the Alps, and all three are mapped by the same reference product. Xavier and Friederike struck ten weeks apart, on either side of the calendar year boundary, but both before the 2018 growing season, so both should peak in 2018. Vaia struck in the same calendar year as Friederike but after its growing season, so it should peak in 2019. Most held-out tiles peak in the predicted year (Fig.~\ref{fig:clock}b). 

\subsection{Features record severity of an event but not its cause}
\label{sec:boundaries}

\begin{figure}[htbp]
\centering
\includegraphics[width=\textwidth,keepaspectratio]{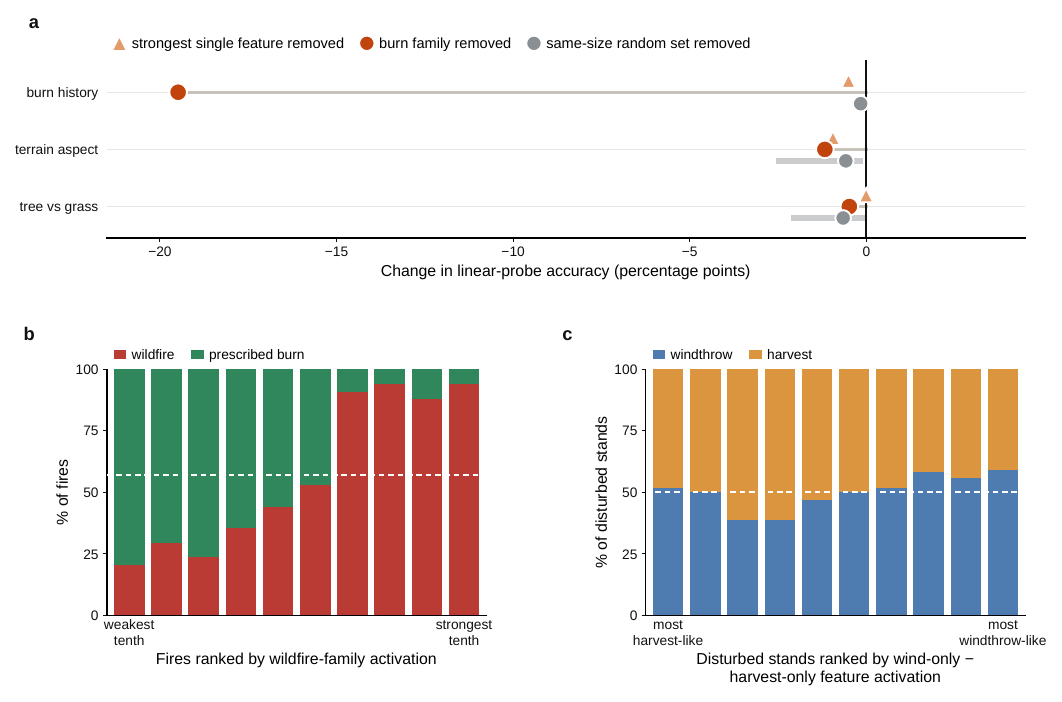}
\caption{\textbf{Removing the burn family removes what a probe uses to detect fire, and features record severity of an event but not its cause.} \textbf{a}, Change in linear probe accuracy when features are removed from the reconstruction. Removing the 32-feature burn family costs the burn probe 19 points, while a single strong feature moves the accuracy only a little, and a matched random set of the same size (mean over 200 draws, band 95\% CI) has nearly no effect. Neither ablation moves other probes (terrain aspect, tree vs grass) by more than random chance. \textbf{b}, All size- and region-matched SE US fires (191 wildfire, 145 prescribed burn) ranked by the burned minus unburned activation of the SE wildfire family and split into deciles. Dashed line is pooled wildfire share (57\%) as baseline if activation carried no differentiating information. The four strongest tenths are 88--94\% wildfire. \textbf{c}, 309 matched pairs of wind and bark-beetle and harvest tile-years, ranked by wind-only minus harvest-only feature activation. Every tenth sits near the pooled 50\%, so the specific features do not resolve cause where the surface signals are similar.}
\label{fig:boundaries}
\end{figure}

A feature family that tracks fire could be how the embedding represents fire, or merely a set of features that correlates with it. To tell these apart, we removed the fire family’s contribution from the Dixie fire embeddings and re-evaluated a linear probe trained to find the burn. Its accuracy fell by 19 percentage points, whereas removing the same number of random features had no measurable effect (Fig.~\ref{fig:boundaries}a). The loss was specific to fire since the accuracy for terrain aspect and for tree versus grass cover were unaffected. The direction a linear probe uses to detect fire therefore lies largely within the fire family, which supports the family being how the embedding represents fire rather than a correlate of it.

Features can represent disturbance severity, but cannot distinguish between events that have different causes but result in the same land surface condition. For example, prescribed burns, the low-intensity understory fires set for management, activate the wildfire family only weakly, enough for a single threshold to separate the two (Fig.~\ref{fig:boundaries}b). They do not separate causes that leave the same surface. Windthrow, bark-beetle damage and harvest all remove the canopy abruptly, and thus their families share most features. Even the features unique to one family respond equally to both (Fig.~\ref{fig:boundaries}c). The embeddings record what a disturbance did to the surface, not what caused it, so they separate causes only when the surfaces differ. Fire does leave a distinct surface, and accordingly the western US burn family responds to European fires but not to European harvest or windthrow (\sinote{note:specificity}).

Two further experiments demonstrate the temporal limits of decomposing annual embeddings. RADD radar alerts \cite{reiche2021RADD} date Amazon clearings to the day. If the embeddings carried timing finer than a year, a clearing's position within the year would predict which annual embedding shows it. We find that across 17{,}014 RADD-dated clearing plots, the two are dated equally well, so the remaining uncertainty comes from annual compositing. We also reversed the screen to look for features whose contrast rises in the years before a mapped disturbance. On held-out units the pre-event effect is small and its interval includes zero (+0.014, CI $-0.001$ to $+0.136$), so we find no early warning signal.

\subsection{Embedding contents differ between models}
\label{sec:crossmodel}

\begin{figure}[htbp]
\centering
\includegraphics[width=\textwidth,keepaspectratio]{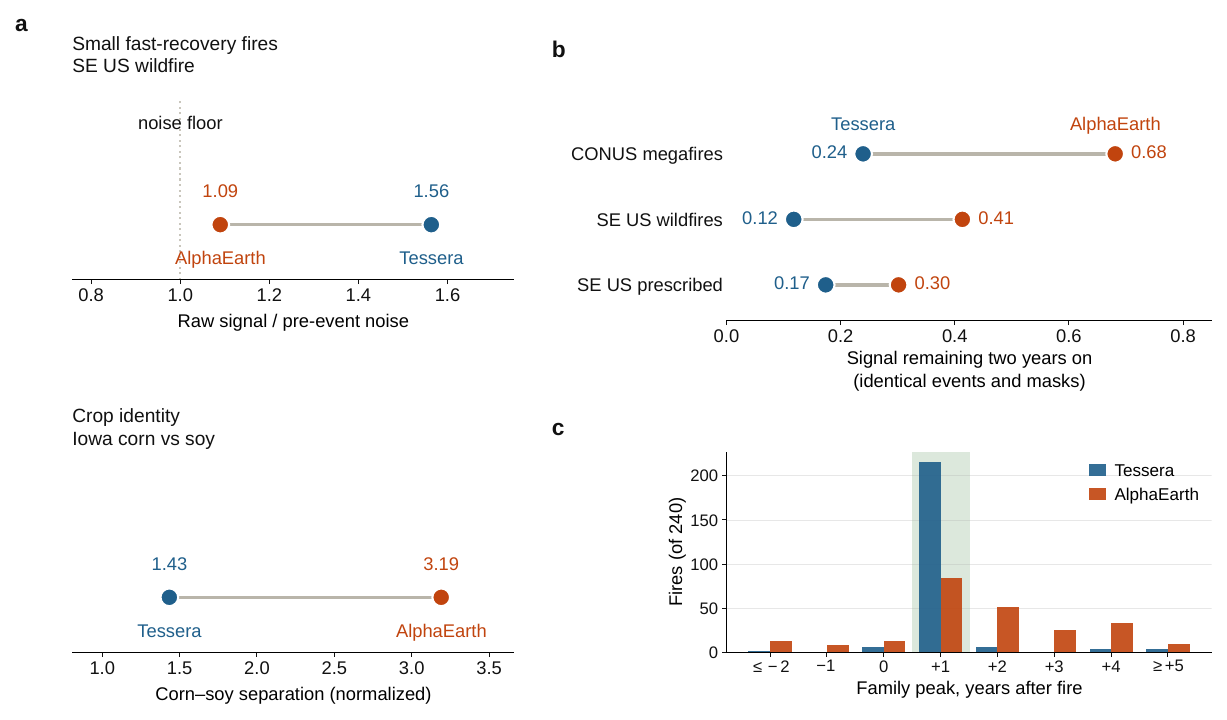}
\caption{\textbf{Tessera and AlphaEarth retain different phenomena on identical events.} Panels use raw embeddings on identical events and masks. \textbf{a}, Top: small fast-recovery fires (SE US wildfire), raw signal over pre-event noise. Tessera sits at 1.56 and AlphaEarth at 1.09, at the noise floor (dotted). The identical feature screen yields 15 features in Tessera and none in AlphaEarth over the same 21 events. Bottom: crop identity (Corn Belt corn vs soy, separation of class means in units of within-class spread), Tessera 1.43 and AlphaEarth 3.19. \textbf{b}, Fraction of disturbance signal remaining two years after identical events, per setting. AlphaEarth retains more in all three (ratios 2.8, 3.5, 1.7). \textbf{c}, Each model's family-peak offset from the fire year over 240 western US megafires both screened. Green band marks the expected window.}
\label{fig:crossmodel}
\end{figure}

Finally, we compare the two products on identical events. Both Tessera and AlphaEarth were trained on largely the same physical observational data of the Earth surface from Sentinel-1 and Sentinel-2. If an embedding's contents were determined by what the satellites observe, the two models should record the same phenomena. They do not, and one possible explanation is that their training objectives treat time differently. Tessera is trained so that different random subsets of one year's observations yield the same embedding, so its embedding must describe what happened in that year. We call it event-indexed. AlphaEarth models each location as a continuous function of time, so each embedding also draws on observations from neighboring times. We hypothesize that this makes it state-indexed, or that it tracks the lasting condition of a place and smooths over brief events. To make it so differences cannot arise from our decompositions, all comparisons in this section use raw embeddings on the same events (Fig.~\ref{fig:crossmodel}).

Each model records something the other does not (Fig.~\ref{fig:crossmodel}a). For small fires that recover quickly, such as southeastern US wildfires whose understory regrows within a year, we asked how much more distinct burned and unburned areas become after the fire than they already were before it. In Tessera the fire stands out (median ratio 1.56) but in AlphaEarth it is only somewhat distinguishable (1.09, where 1 means no change). Screening the same fires that yield the SE wildfire family in Tessera finds no family at all in AlphaEarth, so these fires appear to be weakly identified from its embeddings. Crop identity shows the reverse, where corn and soy are 1.43 within-class standard deviations apart in Tessera and 3.19 in AlphaEarth. Both fit the hypothesis once lasting condition is read on the scale of a growing season. A crop holds a field for the whole season, whereas an understory fire is gone within weeks.

The models also differ in how long they retain a disturbance (Fig.~\ref{fig:crossmodel}b). We measured the disturbance signal two years after its peak as a fraction of the peak, on the same events in both models. AlphaEarth retains 2.8, 3.5 and 1.7 times more of the signal than Tessera for western US megafires, southeastern US wildfires and southeastern US prescribed fires, respectively. Of 240 western US megafires screened in both models, Tessera's family peaks in the first post-fire growing season for 216, and AlphaEarth's for 84, with the remainder spread over the following two to four years (Fig.~\ref{fig:crossmodel}c).

\section{Discussion}
\label{sec:discussion}

We decomposed two annual embedding products into named features and verified them against external reference data. By doing so, we found that annual Earth embeddings are interpretable records of the land surface. They record abrupt canopy change, resolve its severity, and date it to the calendar year containing the first growing season after the disturbance. The findings follow ecological intuition, which is consistent with the decomposition capturing real structure rather than artifacts of the method. Had it produced incoherent features like burn features peaking before fires, or a clean separation of causes that leave identical surface signals, either would have pointed to artifacts of the method. Representations that are already widely used can thus be decomposed into physically interpretable features and checked against independent data.

The two models observe the same planet yet retain different phenomena, which suggests that the training objective may shape what an embedding contains. Benchmarks may not detect this, because they score only the tasks that were included, so a phenomenon that no benchmark covers may be missing from a model. An embedding product that performs well on available benchmarks may therefore be uninformative about the phenomenon a user needs, as our comparison shows for small fires that recover quickly. This matters most where embeddings inform decisions that must be defensible case by case, such as compliance with deforestation regulations, where wall-to-wall map products have often fallen short at the level of individual plots \cite{freitasbeyer2025eudr}. Before building a monitoring or compliance system on an embedding product, users can check whether it contains the target phenomenon at the required temporal resolution. The method used here needs only the embeddings and a reference product for that phenomenon, so it applies to closed products as well as open ones, as our analysis of AlphaEarth's public embeddings shows. Isolating the effect of the objective would require a set of models trained on a fixed Sentinel subsample with different objectives, which is beyond the scope of this study.

The cross-model comparison rests on one pair of models, because Tessera and AlphaEarth are the only two with publicly available global embeddings, so the link between objective and retention is correlational. Errors in the reference products mostly reduce agreement, and as such our validation rates likely understate how faithfully the embeddings record these events. This assumes the reference errors are largely independent of the embeddings, which may not hold where both derive from similar satellite data. We verify features at the level of individual events and their severity. Finer distinctions, such as causal attribution, timing within a year and boundaries between severity classes, remain open. Even so, sparse decomposition with external verification offers a practical way to probe into what Earth observation foundation models contain.

\section{Methods}
\label{sec:methods}

\subsection{Embedding products}
\label{sec:methods-embeddings}

Tessera \cite{tessera2025} annual embeddings (128-dimensional, 10m) were obtained from the public GeoTessera v1 archive. AlphaEarth Foundations \cite{alphaearth2025} annual embeddings (64-dimensional, 10m) were obtained through Google. Cross-model comparisons use variance-matched scaling of AlphaEarth embeddings to Tessera's scale (\sinote{note:cross-model}), and training and screening use the same scale. All analyses are done on a 0.1$^\circ$ tile grid ($\sim$11 km at the equator), which is the native spatial index of the embedding products. Reference data are reprojected onto this grid before any analysis. We use frozen embeddings, and no model was finetuned. 

\subsection{Sparse dictionaries}
\label{sec:methods-sae}

We generate dictionaries with TopK sparse autoencoders \cite{gao2024scaling}, which is a linear encoder over centered embeddings followed by a TopK activation ($k$ active features per embedding) and a linear decoder, trained to reconstruct the embedding under mean-squared error, with periodic reinitialization of dead features. The main configuration is $m=1{,}024$, $k=16$ on Tessera ($8\times$ expansion) and $m=512$, $k=8$ on AlphaEarth (matched $8\times$ expansion). 

One dictionary is trained per region on that region's own tiles and shared by the settings in that region (\sinote{note:dict}). Where a family is applied to new regions (Fig.~\ref{fig:decompose}f), the dictionary is unchanged. Training corpora are embeddings only, shuffled across tiles, with held out reconstruction checked on fresh samples. No labels, reference data products or event dates are used in decomposition. Dictionaries reconstruct held out embeddings with median $R^2$ of 0.9, and fewer than 3\% of features are inactive on average.

\subsection{Feature screening}
\label{sec:methods-screen}

For each unit and year, we compute each feature's mean activation over disturbed pixels minus its mean over undisturbed pixels in the same tile, indexed by years since the event. A family is screened on the median of these contrasts across the screening units. A feature is kept if its contrast in the setting's expected window (see Sec.~\ref{sec:methods-verdicts}) exceeds a threshold $c$, its absolute contrast in every earlier year is below $c/2$, and its contrast from the third year after the event onward is below half its in-window value. Because median contrast falls roughly 50-fold across the dictionary sizes we swept, $c$ is set relative to each dictionary, as a multiple of the median in-window contrast over active features, and is held fixed across dictionaries, models and settings. Varying the multiplier from a quarter to four times its value changes family sizes, but every setting remains above its null and no interpretation changes (\sinote{note:screen}, Table~\ref{tab:anchor}).

\subsection{Validation corpus and scoring}
\label{sec:methods-verdicts}

A unit is either one mapped event or one tile-year, defined as the pixels of one disturbance type disturbed in one year in a 0.1$^\circ$ tile. Units are selected from reference data alone (\sinote{note:selection}). Event-dated products (MTBS, NBAC, Geoscience Australia, FORWIND) record when a disturbance occurred, whereas detection-dated products (EFDA, Hansen) record the first annual composite in which it is visible, which can be a year later. Each setting's expected window is fixed before screening. We use the year after the event, $\{Y{+}1\}$, where the reference is event-dated and the burned vegetation does not regrow within the growing season in which it burns (western US fire, boreal fire), so the first growing season showing the burned surface is the following year. Otherwise we use the event year or the year after, $\{Y, Y{+}1\}$. 

Each family is screened on a subset of its setting's units and scored only on the held-out remainder (\sinote{note:selection}). A held-out unit is validated when its family's mean activation contrast is highest inside the expected window, and otherwise it peaks early or late. Crop tiles are held out in time, screened on 2018--2021 and scored on 2022--2024. Crop units are tiles, scored by classification rather than dating. There, a tile validates when its held-out classification accuracy reaches 0.70 in the crop atlas (seven competing classes) or 0.80 in crop rotation (corn vs. soy). A setting's validation rate is the share of its held-out units that validate.

\subsection{Task nulls and uncertainty}
\label{sec:methods-stats}

For a disturbance setting, the task null is computed by drawing a random feature set of the family's size from the same dictionary, scoring the same held out units under the same expected window, repeated for 200 draws. Then we calculate the null median, 95\% interval and $P(\mathrm{null} \ge \mathrm{observed})$. Because outcomes cluster within tiles, all confidence intervals on validation rates and rate differences are tile-clustered bootstraps over 10{,}000 resamples of tiles. For probes, we use logistic heads on frozen embeddings or reconstructions with held-out evaluation. Ablation subtracts a family's contribution from each embedding and compares against a matched random set of equal size. The cropland data layers analysis does not have a null, since crop units are scored by held-out classification accuracy, for which a peak-offset null is not defined. Ablation methods are described in detail in \sinote{note:causal}.

\subsection{Reference data products}
\label{sec:methods-data}

Monitoring Trends in Burn Severity (MTBS) fire perimeters and ignition dates (western and southeastern US, including the prescribed-fire type field) \cite{picotte2020MTBS, eidenshink2007MTBS}, National Burned Area Composite (NBAC, boreal Canada) \cite{hall2020NBAC}, Geoscience Australia Bushfire ignition dates (Australia) \cite{GA2024bushfire}, European Forest Disturbance Atlas (EFDA) disturbance year and agent classes (Europe) \cite{vianasoto2025EFDA, vianasoto2024EFDAdata}, FORWIND storm polygons with event dates \cite{forzieri2020FORWIND}, Hansen Global Forest Change loss year (Amazon, Africa, East Asia) \cite{hansen2013}, rainfall-triggered landslide inventories \cite{berti2025RER2023, ma2024RLBJ, fu2025RLZX}, JRC Global Surface Water yearly water history (new permanent water) \cite{pekel2016GSW}, RAdar for Detecting Deforestation (RADD) alerts (per-plot dating) \cite{reiche2021RADD} from Global Forest Watch, USDA Cropland Data Layer (crop systems) \cite{boryan2011CDL}, and the JRC EU crop map (European winter wheat) \cite{dandrimont2021EUCROPMAP, ghassemi2024EUCROPMAP}. The supplementary additionally uses the global mining land-use dataset \cite{cheng2026mining} and the US Large-Scale Solar Photovoltaic Database (USPVDB) \cite{fujita2023USPVDB}. All data products used are publicly available.

\subsection{Cross-model comparison}
\label{sec:methods-crossmodel}

Cross-model comparisons use raw embeddings on identical events and masks, with AlphaEarth embeddings rescaled to Tessera's variance. For each event and year, we compute the distance between the mean embeddings of disturbed and undisturbed pixels. Signal over noise is the larger of this distance in the event year and the year after, divided by its largest pre-event value. Retention is the signal above the pre-event median two years after its peak, as a fraction of the peak. Corn--soy separation is the distance between class means in units of within-class spread (\sinote{note:cross-model}).

\section*{Supplementary information}
The Supplementary Information follows the reference list.

\section*{Data availability}
All embedding data and reference datasets used are publicly available. Tessera embeddings (\href{https://pypi.org/project/geotessera/}{Geotessera}), AlphaEarth Foundations embeddings (\href{https://developers.google.com/earth-engine/datasets/catalog/GOOGLE_SATELLITE_EMBEDDING_V1_ANNUAL/}{Earth Engine} or \href{https://storage.googleapis.com/alphaearth_foundations/}{GCS}), \href{https://www.mtbs.gov/}{MTBS}, \href{https://cwfis.cfs.nrcan.gc.ca/ha/nfdb?type=nbac}{NBAC}, \href{https://digital.atlas.gov.au/datasets/historical-bushfire-boundaries-version-2-0/about}{Geoscience Australia Bushfire}, \href{https://effis.jrc.ec.europa.eu/}{EFDA}, \href{https://core.ac.uk/outputs/288854916}{FORWIND}, \href{https://storage.googleapis.com/earthenginepartners-hansen/GFC-2023-v1.11/download.html}{Hansen GFC}, \href{https://data.globalforestwatch.org/datasets/gfw::deforestation-alerts-radd/about}{RADD}, \href{https://www.nass.usda.gov/Research_and_Science/Cropland/SARS1a.php}{USDA CDL}.

\bibliographystyle{unsrtnat}
\bibliography{references}

\clearpage
\setcounter{section}{0}
\setcounter{figure}{0}
\setcounter{table}{0}
\setcounter{equation}{0}
\renewcommand{\thesection}{S\arabic{section}}
\renewcommand{\thefigure}{S\arabic{figure}}
\renewcommand{\thetable}{S\arabic{table}}
\renewcommand{\theequation}{S\arabic{equation}}
\renewcommand{\theHsection}{S\arabic{section}}
\renewcommand{\theHfigure}{S\arabic{figure}}
\renewcommand{\theHtable}{S\arabic{table}}
\renewcommand{\theHequation}{S\arabic{equation}}
\makeatletter
\def\sectionname{Supplementary Note}
\def\@seccntformat#1{\@ifundefined{#1name}{}{\csname #1name\endcsname\ }\csname the#1\endcsname\hskip.5em}
\makeatother

\begin{center}
{\LARGE\bfseries Supplementary Information}\\[0.8em]
{\large Earth embeddings are decomposable records of the land surface}
\end{center}
\vspace{1em}
\section{Reference data products}
\label{note:data}

We use public reference data products for our ground-truth dates, perimeters, and class labels:

\begin{enumerate}[label=\arabic*.]
    \item Event-dated products (MTBS July 15 2026 release \citep{picotte2020MTBS, eidenshink2007MTBS}, NBAC \citep{hall2020NBAC}, Geoscience Australia's Historic Bushfire Boundaries \citep{GA2024bushfire}, FORWIND v2 \citep{forzieri2020FORWIND}, and three rainfall-triggered landslide inventories: RER2023, Emilia-Romagna, May 2023 \citep{berti2025RER2023}; RLBJ, Beijing, July 2023 \citep{ma2024RLBJ}; and RLZX, Zixing, July 2024 \citep{fu2025RLZX}) record disturbance occurrence dates.
    \item Detection-dated products (European Forest Disturbance Atlas \citep{vianasoto2025EFDA, vianasoto2024EFDAdata}, Hansen Global Forest Change GFC-2023-v1.11 \citep{hansen2013}, and the JRC Global Surface Water yearly water history v1.4 \citep{pekel2016GSW}) record the first annual composite in which a disturbance or new water is visible. RADD alerts v20260802 \citep{reiche2021RADD} date each radar detection to the day.
    \item Class labels: the USDA Cropland Data Layer \citep{boryan2011CDL} for the crop atlas, crop rotation, and US rice and winter wheat in the invariance test; the JRC EU crop map, 2018 \citep{dandrimont2021EUCROPMAP} and 2022 \citep{ghassemi2024EUCROPMAP} editions, for European winter wheat in the invariance test (Supplementary Note~\ref{note:specificity}); and, for the non-vegetation surfaces (Supplementary Note~\ref{note:detection}), the global mining land-use dataset \citep{cheng2026mining} and the US Large-Scale Solar Photovoltaic Database (USPVDB) v4.0 \citep{fujita2023USPVDB}.
    \item A composited dNBR burn severity layer calculated from Sentinel-2 \citep{drusch2012sentinel2} accessed via Microsoft Planetary Computer. 
    \item We additionally use ESA WorldCover v2 \citep{zanaga2022WorldCover} and Copernicus DEM GLO-30 \citep{copernicus2021GLO30} to get co-resident probe targets for ablation controls, and in the landslide setting to exclude built-up land, water and snow from the undisturbed pixels and to set their slope floor.
\end{enumerate}

We reproject rasters onto the embedding tile grid at acquisition to avoid post-masking resampling. The two date types can differ by up to a year for the same physical event, and storm Vaia illustrates this difference, carrying both a FORWIND event date and EFDA labels. Offset statistics specify their date type, and offsets are not pooled across types. Expected windows are set per setting based on the date type and local growing season, fixed before screening (Section~\ref{sec:methods-verdicts}).

For \mainfig{fig:decompose}, pre- and post-fire median NBR composites are built from cloud-masked Sentinel-2 L2A scenes (SCL clear classes only) over fire-specific windows, and burned is defined as $\Delta$NBR $> 0.27$. The composite covers 69.4\% of the Dixie tile. The AUC in \mainfig{fig:decompose}e is computed over covered pixels.

\section{Embeddings and dictionaries}
\label{note:dict}

\textbf{Embeddings.} Tessera annual embeddings (128-d, 10m) \citep{tessera2025} are read from the public GeoTessera v1 archive on a $0.1^{\circ}$ tile grid. AlphaEarth embeddings (64-d, 10m) \citep{alphaearth2025} are reprojected onto the same grid. All work uses frozen embeddings without finetuning.

\textbf{Dictionary.} We use a TopK sparse autoencoder (SAE) \citep{gao2024scaling} with tied-initialization linear encoder and decoder:
\begin{equation}
\label{eq:sae}
    f = \mathrm{TopK}_k\!\left(\mathrm{ReLU}\!\left(W_e (x - b_{\mathrm{dec}}) + b_e\right)\right), \quad \hat{x} = W_d f + b_{\mathrm{dec}},
\end{equation}
where $W_e \in \mathbb{R}^{m \times d}$, $W_d \in \mathbb{R}^{d \times m}$, and $\mathrm{TopK}_k$ keeps the $k$ largest preactivations and zeroes the rest. The objective is mean squared reconstruction error. Sparsity is imposed architecturally via the TopK rather than by an $L_1$ penalty. Training uses Adam optimizer with learning rate $10^{-3}$, batch size of 4{,}096, 40 epochs and decoder renormalization at each step.

Latents that have not activated within a window are reinitialized to prevent reduction of effective capacity. The corpus is a spatial sample of pixel-years from the setting's own tiles. No label, reference product, or event date is used in dictionary training.

\begin{figure}[htbp]
\centering
\includegraphics[width=0.48\textwidth,keepaspectratio]{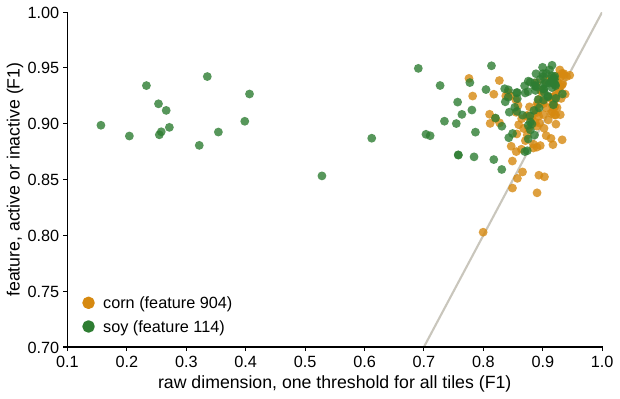}
\caption{\textbf{One fixed feature against one fixed raw-dimension rule.} For each of 100 Corn Belt tiles, the F1 with which the corn and soy features of \mainfig{fig:sparsity}, read as active or inactive, identify their crop (vertical), against the best single raw embedding dimension with one end and threshold chosen on the tile of \mainfig{fig:sparsity} (horizontal). Points above the line are tiles where the feature does better.}
\label{fig:purity}
\end{figure}

\textbf{Feature purity at scale.} \mainfig{fig:sparsity} shows one corn and one soy feature on one Iowa tile. Across the 100 Corn Belt rotation tiles, these are the best single corn and soy features of the Iowa dictionary on 99 and 96 tiles. Read as active or inactive, they identify corn and soy with median F1 0.91 and 0.92. A raw embedding dimension reaches similar accuracy only when its end and threshold are tuned on each tile (0.92 and 0.90). Given a single end and threshold chosen on the tile of \mainfig{fig:sparsity} and applied to every tile, as the features' active-or-inactive reading effectively is, the best raw dimensions reach 0.90 and 0.87, the features do better on 71 and 98 of the 100 tiles, and the soy threshold falls below F1 0.7 on 16 tiles where the soy feature does not (Fig.~\ref{fig:purity}). Raw dimensions also have no fixed reading. Across the crop atlas, 99\% of them have their two 2\% extremes dominated by different land-cover classes.

Reference configurations use an $8\times$ expansion factor for both Tessera ($m=1{,}024$, $k=16$) and AlphaEarth ($m=512$, $k=8$). Dictionaries are trained per region on that region's own tiles. Settings in the same region share a dictionary. The three Europe atlas settings use one pan-European dictionary, and the western US setting uses the dictionary trained on the Dixie tile. Families are scored only through the dictionary they were screened on, which is held fixed when applied to new regions (Figure~\ref{fig:decompose}f).

\section{Measurement and screening}
\label{note:screen}

\textbf{Units and groups.} A unit is one mapped event or one tile-year (the pixels of one disturbance type disturbed in one year within one tile). For each unit and year $y$, the reference product supplies:

\begin{itemize}
    \item Disturbed pixels $D$: pixels the reference product marks as disturbed in the unit's year
    \item Undisturbed pixels $U$: pixels in the same tile not marked as disturbed,(Supplementary Note~\ref{note:selection})
\end{itemize}

Units are dropped if either set falls below the setting's floor, to ensure robust group means (floors are listed per setting in Supplementary Note~\ref{note:selection}).  

\textbf{Contrast.} For feature $j$ and year $y$, the contrast is:
\begin{equation}
\label{eq:contrast}
c_j(y) = \frac{1}{|D|}\sum_{p \in D} f_j(p, y) \;-\; \frac{1}{|U|}\sum_{p \in U} f_j(p, y).\end{equation}
Contrasts are indexed by the offset $o = y - y_{\mathrm{event}}$ throughout, so a calendar year is converted once, at measurement.

\textbf{Aggregation.} Per-unit contrasts are combined into one trajectory per setting as the median over units of the difference. Offsets supported by fewer than three units are dropped where offsets are ragged, as they are when units are dated events rather than tile-years.

\textbf{Anchored bar.} Let $\mathrm{inwin}_j = \max_{o \in W} c_j(o)$, where $W$ is the setting's expected window ($\{Y{+}1\}$ or $\{Y, Y{+}1\}$, Table~\ref{tab:corpus}), so a family is screened on the same years in which its held-out units are scored. The screening threshold is:
\begin{equation}
\label{eq:bar}
    c^{*} = 29 \times \operatorname{median}_{j \in A} \left| \mathrm{inwin}_j \right|,
\end{equation}
where $A$ is the set of features active anywhere in the trajectory. The multiplier is fixed once and reused for every dictionary, model, and setting. An absolute threshold is not comparable across dictionary capacities since the median contrast falls roughly 50-fold across the capacity range studied here, so a fixed bar tightens from $13.8\times$ to $660\times$ the typical contrast as dictionaries grow, which is the artifact in Supplementary Figure~\ref{fig:capacity}.

\textbf{Conditions.} Feature $j$ joins the family if:
\begin{align}
    \mathrm{inwin}_j &> c^{*} & &\text{(large in the expected window)} \label{eq:cond1}\\
    \max_{o<0} \left| c_j(o) \right| &< c^{*}/2 & &\text{(inactive before the event)} \label{eq:cond2}\\
    \max_{o \ge 3} c_j(o) &< \mathrm{inwin}_j / 2 & &\text{(decayed afterwards)} \label{eq:cond3}
\end{align}
An absent offset imposes no constraint, and a missing in-window offset scores zero, so a feature cannot pass on an offset that was never measured.

\textbf{Peak-year outcome.} For a held-out unit with family $F$, the peak offset is $$\hat{o} = \arg\max_{o} \frac{1}{|F|}\sum_{j \in F} c_j(o)$$ and the outcome is $$\hat{o} \in W$$ A setting's validation rate is the mean outcome over its held-out units. Expected windows are fixed per setting before screening and are listed in Table~\ref{tab:corpus}: $\{Y{+}1\}$ where the reference is event-dated and the burned vegetation does not regrow within the growing season in which it burns (western US and boreal fire), $\{Y, Y{+}1\}$ elsewhere.

\textbf{Multiplier sensitivity.} The multiplier in Equation~\eqref{eq:bar} is fixed once, and the corpus does not depend on the choice. Re-deriving every family from its screening units with the multiplier set to 7.25, 14.5, 58 and 116 (which is $\tfrac14\times$, $\tfrac12\times$, $2\times$ and $4\times$ of the $29\times$ multiplier that is used for the main results), and rescoring the same held out units under the unchanged expected window, leaves each setting's validation rate within three points of the reported value in most disturbance settings (Supplementary Table~\ref{tab:anchor}). Family sizes respond strongly to the bar as a threshold should, but in the end the outcomes do not.

\textbf{Seed sensitivity.} Retraining the Dixie dictionary with 5 seeds and re-screening with the same rule gives families of 25--38 features whose western US validation rates range from 0.84 to 0.92, each far above its random-family null (medians 0.12--0.15). Individual features differ between seeds (median best-match decoder cosine 0.63), but each family's mean decoder direction is nearly the same (cosine 0.93--0.96).

\begin{table}[ht]
\centering
\caption{Sensitivity of the held-out validation rate to the screening multiplier. For each disturbance setting the family is rederived from its screening tile-years with the multiplier in Supplementary Equation~\eqref{eq:bar} set to the column's value, and the same held-out units are re-scored under the unchanged expected window. Cells give the validation rate with the family size in parentheses}
\label{tab:anchor}
\small
\setlength{\tabcolsep}{4pt}
\begin{tabular}{@{}lrlllll@{}}
\toprule
Setting & $n$ & 7.25 & 14.5 & \textbf{29} & 58 & 116 \\
\midrule
Western US fire & 207 & 0.884 (74) & 0.894 (62) & 0.889 (32) & 0.894 (11) & 0.841 (4) \\
SE US wildfire & 154 & 0.701 (14) & 0.773 (15) & 0.792 (15) & 0.896 (15) & 0.890 (14) \\
SE US prescribed fire & 109 & 0.761 (8) & 0.752 (11) & 0.798 (12) & 0.817 (10) & 0.798 (11) \\
Boreal fire & 309 & 0.961 (20) & 0.955 (19) & 0.939 (20) & 0.955 (17) & 0.948 (15) \\
SE Australia fire & 274 & 0.661 (37) & 0.664 (33) & 0.624 (30) & 0.639 (26) & 0.631 (20) \\
Europe harvest & 826 & 0.821 (67) & 0.818 (67) & 0.832 (66) & 0.809 (54) & 0.825 (42) \\
Europe wind/bark beetle & 352 & 0.815 (71) & 0.827 (64) & 0.841 (56) & 0.867 (47) & 0.858 (35) \\
Europe fire & 151 & 0.669 (37) & 0.695 (36) & 0.669 (36) & 0.662 (33) & 0.656 (29) \\
Windthrow (Friederike, Xavier, Vaia) & 89 & 0.753 (91) & 0.719 (78) & 0.786 (61) & 0.888 (45) & 0.910 (26) \\
Amazon arc clearing & 546 & 0.941 (35) & 0.934 (33) & 0.918 (32) & 0.929 (30) & 0.919 (29) \\
Africa forest loss & 140 & 0.800 (64) & 0.893 (50) & 0.879 (35) & 0.879 (24) & 0.893 (11) \\
East Asia forest loss & 317 & 0.965 (87) & 0.972 (69) & 0.978 (45) & 0.972 (24) & 0.969 (8) \\
Rainfall landslides (Italy, China) & 71 & 0.958 (55) & 0.986 (40) & 1.000 (23) & 1.000 (9) & 1.000 (2) \\
New permanent water & 158 & 0.823 (27) & 0.848 (25) & 0.867 (19) & 0.823 (16) & 0.861 (12) \\
\bottomrule
\end{tabular}
\end{table}

\section{Event corpus results}
\label{note:corpus}

Table~\ref{tab:corpus} gives every setting's reference data product, date type, expected window, held-out unit count, validation rate, and null baseline. The crop rows have no nulls because crop units are scored by held-out classification accuracy, for which a peak-offset null is undefined.

\begin{table}[ht]
\centering
\caption{Full results over 16 settings, 3{,}940 reference units. Validation is the share of held-out units whose family peak falls in the expected window (disturbance settings) or held-out classification accuracy (crop settings). Nulls are same-size random-family draws (median and 95\% interval over 200 draws)}
\label{tab:corpus}
\small
\setlength{\tabcolsep}{4pt}
\resizebox{\textwidth}{!}{\begin{tabular}{@{}lllllrrl@{}}
\toprule
Setting & Product & Embeddings & Dates & Window & $n$ & Validated & Null (median [95\% int.]) \\
\midrule
Western US fire & MTBS & Tessera & event-dated & \{Y$+$1\} & 207 & 0.889 & 0.140 [0.024, 0.406] \\
SE US wildfire & MTBS & Tessera & event-dated & \{Y, Y$+$1\} & 154 & 0.792 & 0.312 [0.155, 0.656] \\
SE US prescribed fire & MTBS & Tessera & event-dated & \{Y, Y$+$1\} & 109 & 0.798 & 0.248 [0.101, 0.505] \\
Boreal fire & NBAC & AlphaEarth & event-dated & \{Y$+$1\} & 309 & 0.939 & 0.196 [0.042, 0.523] \\
SE Australia fire & GA Bushfire & AlphaEarth & event-dated & \{Y, Y$+$1\} & 274 & 0.624 & 0.369 [0.201, 0.639] \\
Europe harvest & EFDA & Tessera & detection-dated & \{Y, Y$+$1\} & 826 & 0.832 & 0.301 [0.113, 0.529] \\
Europe wind/bark beetle & EFDA & Tessera & detection-dated & \{Y, Y$+$1\} & 352 & 0.841 & 0.254 [0.062, 0.514] \\
Europe fire & EFDA & Tessera & detection-dated & \{Y, Y$+$1\} & 151 & 0.669 & 0.311 [0.159, 0.543] \\
Windthrow & FORWIND & Tessera & event-dated & \{Y, Y$+$1\} & 89 & 0.786 & 0.202 [0.022, 0.697] \\
Amazon arc clearing & Hansen GFC & AlphaEarth & detection-dated & \{Y, Y$+$1\} & 546 & 0.918 & 0.319 [0.091, 0.665] \\
Africa forest loss & Hansen GFC & Tessera & detection-dated & \{Y, Y$+$1\} & 140 & 0.879 & 0.357 [0.107, 0.700] \\
East Asia forest loss & Hansen GFC & Tessera & detection-dated & \{Y, Y$+$1\} & 317 & 0.978 & 0.457 [0.047, 0.868] \\
Rainfall landslides & RER2023, RLBJ, RLZX & Tessera & event-dated & \{Y, Y$+$1\} & 71 & 1.000 & 0.366 [0.056, 0.818] \\
New permanent water & JRC GSW & Tessera & detection-dated & \{Y, Y$+$1\} & 158 & 0.867 & 0.222 [0.057, 0.595] \\
Crop atlas & USDA CDL & Tessera & --- & --- & 102 & 0.824 & --- \\
Crop rotation & USDA CDL & Tessera & --- & --- & 100 & 0.950 & --- \\
\bottomrule
\end{tabular}}
\end{table}

Restricting random draws to live features outside the family lowers the null in all but one setting, where it moves by 0.01 (Table~\ref{tab:null_live}), so the nulls in Table~\ref{tab:corpus} are conservative.

\begin{table}[h]
\centering
\caption{\textbf{Random family null is conservative.} Null validation rate (median over 200 draws) as implemented, with random feature sets drawn from the whole dictionary including family features, and with draws restricted to live features outside the family, against the observed rate. A feature is live if its contrast is nonzero in at least one held-out unit; dead features are 0.0--3.5\% of each dictionary. Restricting draws lowers the null in 12 of 14 settings.}
\label{tab:null_live}
\small
\begin{tabular}{lrrr}
\toprule
Setting & Null (as implemented) & Null (live, non-family) & Observed \\
\midrule
Western US fire & 0.14 & 0.09 & 0.89 \\
SE US wildfire & 0.31 & 0.30 & 0.79 \\
SE US prescribed fire & 0.25 & 0.26 & 0.80 \\
Boreal fire & 0.20 & 0.17 & 0.94 \\
SE Australia fire & 0.37 & 0.32 & 0.62 \\
Europe harvest & 0.30 & 0.19 & 0.83 \\
Europe wind/bark beetle & 0.25 & 0.18 & 0.84 \\
Europe fire & 0.31 & 0.28 & 0.67 \\
Windthrow (Friederike, Xavier, Vaia) & 0.20 & 0.12 & 0.79 \\
Amazon arc clearing & 0.32 & 0.24 & 0.92 \\
Africa forest loss & 0.36 & 0.34 & 0.88 \\
East Asia forest loss & 0.46 & 0.30 & 0.98 \\
Rainfall landslides (Italy, China) & 0.37 & 0.31 & 1.00 \\
New permanent water & 0.22 & 0.22 & 0.87 \\
\bottomrule
\end{tabular}
\end{table}

\section{Unit selection and screening sets}
\label{note:selection}

For each setting we give the selection rule and the minimum group sizes below which a unit is dropped. All tiles are $0.1^{\circ}$ grid cells.

\textbf{Western US fire.} We use every MTBS wildfire or fire complex of at least 20{,}000 acres in the western CONUS, 2018--2023, whose perimeter covers at least 15\% of one tile, which gets us 242 fires. The burn mask is the MTBS perimeter rasterized on the tile grid, and the undisturbed set is every other pixel in the tile. MTBS perimeters contain unburned islands, so the contrasts are conservative. A fire is scored only if every year has at least 300 burned and 300 undisturbed pixels at the screening stride, which leaves 207 of the 242 fires. The feature family is the 32-feature burn family screened on the Dixie fire (\mainfig{fig:decompose}), which is not among the scored fires.

\textbf{Southeastern US wildfire and prescribed fire.} We filtered for MTBS events where both types are drawn from the 1{,}000--5{,}000 acre band, both are confined to $-100$ to $-80^{\circ}$E, 24--40$^{\circ}$N, and an event qualifies if its perimeter intersects no other MTBS perimeter 2017--2025 (southeastern prescribed fire is typically in a 1--3 year return regime, and this removes 92\% of prescribed candidates). The undisturbed pixels are the tile's never-burned remainder. We use a floor filter of at least 300 burned and 2{,}000 undisturbed pixels. The SE wildfire family is screened on 2018 wildfires and 2019--2023 wildfires are held out. The prescribed-burn family is screened on 2018--2019 prescribed burns, and prescribed burns from 2020--2023 are held out.

\textbf{Boreal fire.} We use every NBAC fire of at least 10{,}000ha, 2018--2023, whose perimeter is dominated by one tile, on AlphaEarth (Tessera did not cover Canada fully for the temporal range, at the time of testing). We use a floor filter of at least 300 burned pixels. The feature family is screened on 2018 fires, and 2019--2023 fires are held out.

\textbf{Southeastern Australia fire.} We use every Geoscience Australia mapped Black Summer wildfire of at least 2{,}000ha. The prescribed comparison is the 80 largest prescribed burns 2019--2021 (at least 500ha) on tiles no wildfire claims, so wildfires and prescribed burns do not share a tile. We use a floor filter of at least 300 burned pixels. The feature family is screened on fires ignited September--October 2019, and all other fires are held out, mostly ignited November 2019 to March 2020 and 14 in July--August 2019.

\textbf{Europe atlas (harvest, wind/bark-beetle, fire).} Tiles are sampled by EFDA disturbance agent density, where a disturbance type must disturb at least 2\% of the tile's forest over 2018--2022, with at most 6 tiles per country per type so that no country dominates, which gets 436 tiles across 36 countries. A tile-year is used if it holds at least 600 sampled disturbed pixels and its tile carries at least 2{,}000 undisturbed forest pixels. One Europe-wide dictionary is trained on all tiles, with one family per disturbance type. Each type's feature family is screened on its 2018 tile-years, with tile-years 2019--2022 held out.

\textbf{Windthrow.} FORWIND damage polygons are aggregated to tiles, and for each of three storms we use every tile holding at least 25ha of mapped damage, up to the 60 largest. This results in 35 Friederike, 6 Xavier and 60 Vaia tiles. Additionally tiles need at least 300 damaged and 2{,}000 undisturbed forest pixels. One windthrow family is screened on 12 randomly chosen Friederike tiles and scored on the remaining 23 Friederike tiles and on every Xavier and Vaia tile, none of which are used for screening. The three storms are pooled as one setting for validation, and reported per storm for timing (\mainfig{fig:clock}b).

\textbf{Amazon arc clearing.} Tiles are sampled from Hansen GFC with at least 3\% of tile forest lost 2018--2022, ranked by clearing pixels, with at least 20{,}000 forest and 20{,}000 intact pixels, spanning 75.2--43.4$^{\circ}$W. Tile-year floors as in the Europe atlas. The Amazon clearing family is screened on 2018 clearing tile-years with tile-years 2019--2022 held out.

\textbf{Africa and East Asia forest loss.} We use the same reference product and test as the Amazon arc, on Tessera, in the 2.5$^{\circ}$ blocks where Tessera covers 2017--2022. Forest is tree cover of at least 60\% in 2000. A unit is a tile-year, or the forest pixels of a tile lost in one year 2018--2023, when they cover at least 25ha, against forest on the same tile never lost. Tiles are ranked by 2018--2023 loss, with at most 4 per block and 80 per region. We use a floor filter of at least 300 lost and 2{,}000 undisturbed pixels. One dictionary is trained per region, its family is screened on the region's 2018 tile-years, and tile-years 2019--2023 are held out.

\textbf{Rainfall landslides.} We used three inventories of landslides mapped by hand from high-resolution imagery after one rainfall event each, from RER2023 (Emilia-Romagna, Italy, May 2023; 80{,}997 polygons on 0.2m aerial photographs) \cite{berti2025RER2023}, RLBJ (western Beijing, China, late July 2023; 15{,}383 polygons, field checked) \cite{ma2024RLBJ} and RLZX (Zixing, Hunan, China, July 2024; 19{,}403 polygons on 2m imagery, checked against drone surveys) \cite{fu2025RLZX}. We use every tile holding at least 15ha of an event's mapped landslides (by polygon centroid). The undisturbed pixels are land more than 50m from any mapped landslide and at least as steep as the 25th percentile of the landslide pixels' slope (Copernicus GLO-30), so the contrast is not merely steep against flat ground. Built-up, water and snow (ESA WorldCover) are excluded. We use a floor filter of at least 300 landslide and 2{,}000 undisturbed pixels. One dictionary is trained on all three events' tiles. As for windthrow, the family is screened on 15 randomly chosen Emilia-Romagna tiles and scored on the remaining Emilia-Romagna tiles and on every Beijing and Zixing tile, none of which are used for screening.

\textbf{New permanent water.} Using the JRC Global Surface Water yearly history (Landsat) \cite{pekel2016GSW}, a 30m pixel has onset year $Y$ (2018--2021) if it is land in each of the three years before $Y$, water in $Y$ and every following year to 2021, and permanent water in 2021. Most such water is reservoirs filling or lakes rising. Candidate tiles are found from the published JRC transition and occurrence layers (new permanent water seen in at most 30\% of 1984--2021 observations) and dated per tile. A unit is one tile on its dominant onset year with at least 25ha of new water, up to 100 tiles per onset year. The new-water pixels are eroded by one pixel against the 30m label edges, and the undisturbed pixels are land that was water in no year 2015--2021 and lies more than 60m from any pixel that was. We use a floor filter of at least 300 new-water and 2{,}000 undisturbed pixels. The family is screened on 2018 tiles, with offsets held by fewer than 3 units dropped, and onsets 2019--2021 are held out.

\textbf{RADD per-plot dating.} Over the Amazon arc tile set, we segment high-confidence RADD alert pixels (confidence class 3) with alert dates 2020-01-01 to 2023-12-31 into connected-component plots. A plot's clearing date is the median alert date of its pixels, and its predicted year is the argmax over 2017--2025 of the plot-mean activation of the Amazon clearing family. This yields 17{,}014 plots over 140 tiles. Every RADD plot post-dates the Amazon clearing family's 2018 screen, so all are held out in time.

\textbf{Crop atlas.} We select 102 tiles from CDL class density alone, measured on the 2018 and 2020 layers, for corn/soy (Corn Belt), winter wheat (Kansas/Oklahoma), rice (Arkansas and Sacramento Valley) and cotton (Texas high plains), and for two further classes that compete in the assignment but are not reported on their own, perennial crops (Central Valley) and fallow (High Plains). One national agricultural dictionary is trained on all tiles. Per-class families are screened on 2018--2021 (one-vs-other-cropland contrast at the anchored bar) and years 2022--2024 are held out. Each held-out pixel-year is assigned the class whose family mean activation is largest, with all seven classes competing at once, and scored against the CDL. Pooled over tiles, held-out accuracy is corn 0.972, soy 0.926, winter wheat 0.795, rice 0.841 and cotton 0.620 (unweighted mean 0.831). A tile validates when its held-out accuracy on its own system's classes reaches 0.70, the bar also used for the pooled class accuracies: corn/soy 18 of 18 tiles, rice 16 of 16, winter wheat 16 of 17, fallow 15 of 16, perennial 13 of 17 and cotton 6 of 18. CDL's own producer accuracy ($\sim$90--95\% for corn/soy, lower for minor classes) creates an upper bound for what is achievable.

\textbf{Crop rotation.} We select the 100 Corn Belt tiles with the highest corn-plus-soy fraction in CDL 2020. Each tile has its own local dictionary, with corn and soy features screened on CDL 2018--2021 and 2022--2024 held out. A tile validates when its held-out corn-versus-soy accuracy from features alone reaches 0.80, as 95 of 100 do.

\section{Families for non-vegetation surfaces}
\label{note:detection}

The dated settings ask whether a family records when a surface changed. Here we ask a simpler question of whether the embedding contains features for a surface that is not vegetation at all.

\textbf{Mining land.} Within hand-digitized mine outlines, the global mining land use dataset classes every pixel by a random forest on Sentinel-2 and TanDEM-X elevation change as open pit, waste dumping site, water body, facility, general disturbed land, bare soil or vegetation \cite{cheng2026mining}. The first five classes are mining land. We draw up to 30 large sites per world region (at least 0.5~km$^2$ of mining land, one site per tile, nine regions), read in the Tessera year closest to the label year and within two years of it. The surroundings are the tile's pixels more than 500m from any labeled polygon.

\textbf{Solar farms.} We use ground mounted farms in the US Large-Scale Solar Photovoltaic Database (USPVDB), digitized from aerial imagery \cite{fujita2023USPVDB}, that were online by 2023, read in Tessera 2024 so that every farm is built. Units are the 300 tiles with the most farm area (at least 25ha, at most 15 per state, 40 states). Panel pixels lie inside a farm outline, eroded by one 20m pixel against mixed edges. The surroundings are pixels more than 100m and at most 2~km from any mapped farm of any year.

\textbf{Method.} One dictionary is trained per setting. The family is screened on a random fifth of the sites, at the anchored bar on the median contrast of target against surroundings, as in the crop atlas. A pixel is assigned to the target when the family's mean activation exceeds one cutoff, set on the screening sites midway between the medians of their target and surroundings activations, and applied unchanged to every held-out site. A held-out site validates when its balanced accuracy (target and surroundings pixels weighted equally) reaches 0.70, same as the crop atlas's bar. The null draws 200 random families of the same size, each with its own cutoff set in the same way. For transfer, the mining family is also screened on four regions (Eastern Europe, Northern America, Oceania, rest of Europe) and scored on the other five, and the solar family is screened on the South and West census regions and scored on the Midwest and Northeast.

\textbf{Results.} Both families validate on most held out sites, far above random families, and transfer across regions (Table~\ref{tab:detection}).

\begin{table}[h]
\centering
\caption{\textbf{Families for non-vegetation surfaces.} Accuracy: median balanced accuracy. AUC: median per-site AUC of the family's activation, target against surroundings.}
\label{tab:detection}
\small
\begin{tabular}{lrrrrrrr}
\toprule
 & Family & Screened / held out & Pass & Null & Accuracy & AUC & Cross-region \\
\midrule
Mining land & 23 & 43 / 174 & 0.82 & 0.07 (0.31) & 0.81 & 0.94 & 0.82 ($n = 98$) \\
Solar farms & 14 & 60 / 239 & 0.75 & 0.03 (0.31) & 0.87 & 0.97 & 0.82 ($n = 66$) \\
\bottomrule
\end{tabular}
\end{table}

\section{Feature counts are instrument-relative}
\label{note:capacity}

\begin{figure}[htbp]
    \centering
    \includegraphics[width=\textwidth,keepaspectratio]{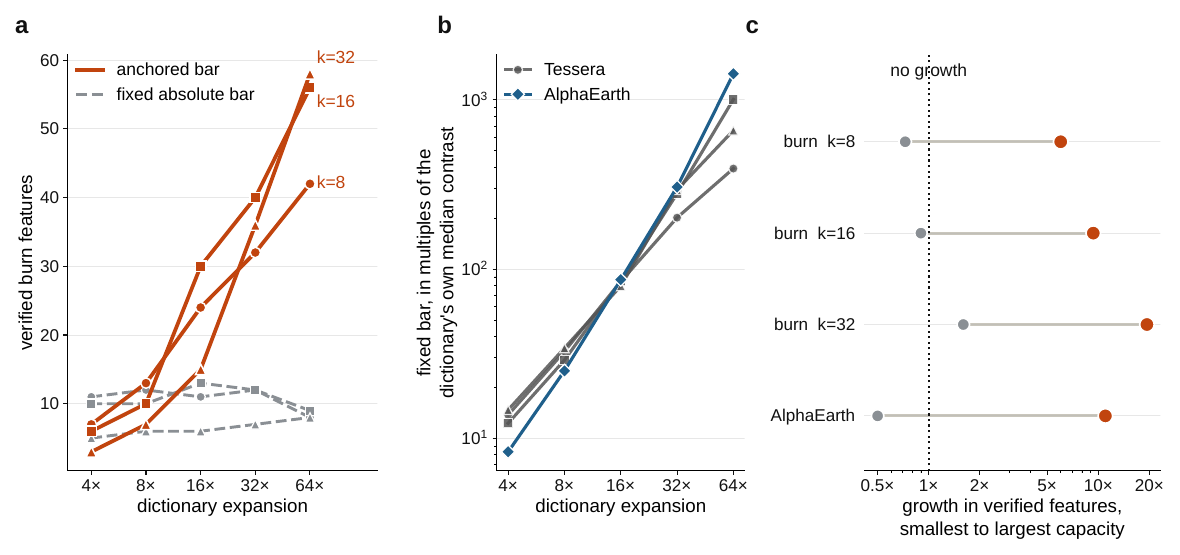}
    \caption{\textbf{Verified feature counts under two screening bars.} \textbf{a}, Externally verified burn-history features against dictionary expansion ($4\times$ to $64\times$ the 128-dimensional embedding), for three sparsity levels $k$, under a bar anchored to each dictionary's contrast distribution (solid) and a fixed absolute bar (dashed), on the same dictionaries with the same verification. \textbf{b}, The fixed bar, in multiples of each dictionary's median in-window contrast, tightens from $13.8\times$ to $660\times$ (Tessera in gray, AlphaEarth in blue). \textbf{c}, Growth ratio (verified features at largest over smallest capacity) for the four measured settings, with the dotted rule marking no growth. The anchored bar shows growth in 4 of 4 settings, and the fixed bar sits at or below 1 in three of the four and reaches 1.6 at $k=32$.}
    \label{fig:capacity}
\end{figure}

Supplementary Figure~\ref{fig:capacity} presents the result summarized in Section~\ref{sec:methods-sae} regarding the number of externally verified features an embedding yields depending on how the screening bar scales with dictionary capacity. Panel a shows verified burn-history features against dictionary expansion under both bars on the same dictionaries. Panel b shows the mechanism, in which the fixed bar, expressed in multiples of each dictionary's own median in-window contrast, tightens from $13.8\times$ to $660\times$ across the sweep. Panel c shows the generality, with growth under the anchored bar in all 4 settings spanning three sparsity levels and both models.

Verification is external throughout because a feature counts only if it peaks in the year after the fire of \mainfig{fig:decompose}f, so the growing count is a count of verified concepts rather than of firing latents. The multiplier in Equation~\eqref{eq:bar} is set so that the anchored and fixed bars coincide at the dictionary configuration where the fixed bar was originally calibrated, and is not changed later.

\section{Cross-model comparability}
\label{note:cross-model}

Comparisons between Tessera (128-d) and AlphaEarth Foundations (64-d) are matched on three axes:

\begin{itemize}
    \item Units: AlphaEarth embeddings are rescaled into Tessera's scale by a single variance-matching factor. Dictionary training and screening operate in the same scale.
    \item Capacity: dictionaries are compared at matched expansion factor, dictionary size over embedding dimension, rather than at matched absolute size.
    \item Pipeline: in the cross-model dissociation (Figure~\ref{fig:crossmodel}), events, masks, screens, and expected windows are identical across models, and the quantities are computed on raw embeddings.
\end{itemize}

The raw-embedding quantities in \mainfig{fig:crossmodel} are defined per unit from
\begin{equation}
d(y) = \left\lVert \overline{x}_{\mathrm{disturbed}}(y) - \overline{x}_{\mathrm{undisturbed}}(y) \right\rVert
\end{equation}
the distance between the mean embeddings of disturbed and undisturbed pixels in year $y$. The signal-to-noise ratio is
\begin{equation}
\frac{\max(d[Y],d[Y{+}1])}{\max_{y<Y} d(y)},
\end{equation}
so 1.0 is the pre-event noise floor. The persistence fraction is
\begin{equation}
\frac{s(\mathrm{peak}{+}2)}{s(\mathrm{peak})},
\quad
s(y) = d(y) - \operatorname{median}_{y<Y} d,
\end{equation}
computed only where the peak signal exceeds a tenth of the noise floor.

\section{Specificity and invariance}
\label{note:specificity}

\textbf{Specificity.} Harvest and windthrow also remove canopy abruptly, so we tested whether the burn family responds to fire specifically or to canopy loss in general. We scored the unchanged 32-feature western US fire family (\mainfig{fig:decompose}) on European fire, harvest and windthrow units. Each tile-year is contrasted against the tile's undisturbed forest as in Equation~\eqref{eq:contrast}, and its peak contrast over 2017--2025 is compared with that of 200 random 32-feature sets from the same dictionary. All fire units are used since the family was not screened on European data.

The western US fire family activates on European fire (median 87th percentile of random sets) but not on harvest or windthrow (45th and 42nd, Table~\ref{tab:specificity}). Its response to European fire is weaker than on the Dixie fire, and 38\% of fire units exceed the random 95th percentile, against 3\% of harvest and none of 89 windthrow units.

\begin{table}[htbp]
\centering
\caption{\textbf{Western US burn family on European disturbances.} The unchanged 32-feature family, scored through the Dixie dictionary. Each unit is a tile-year contrasted against the tile's undisturbed forest; percentiles are relative to 200 random 32-feature sets from the same dictionary, per unit. Window is the share peaking in $\{Y, Y{+}1\}$. Windthrow is the corpus windthrow set (held-out Friederike, Xavier and Vaia tiles).}
\label{tab:specificity}
\small
\begin{tabular}{lrrr}
\toprule
 & European fire & Harvest & Windthrow \\
\midrule
Units & 184 & 69 & 89 \\
Peak contrast, median (max) & 0.059 (0.51) & 0.009 (0.05) & 0.010 (0.06) \\
Random-set peak, median & 0.021 & 0.010 & 0.017 \\
Percentile vs random, median & 87th & 45th & 42nd \\
Above random 95th percentile & 38\% & 3\% & 0\% \\
Peaks in window & 68\% & 36\% & 12\% \\
\bottomrule
\end{tabular}
\end{table}

\begin{figure}[htbp]
\centering
\includegraphics[width=\textwidth,keepaspectratio]{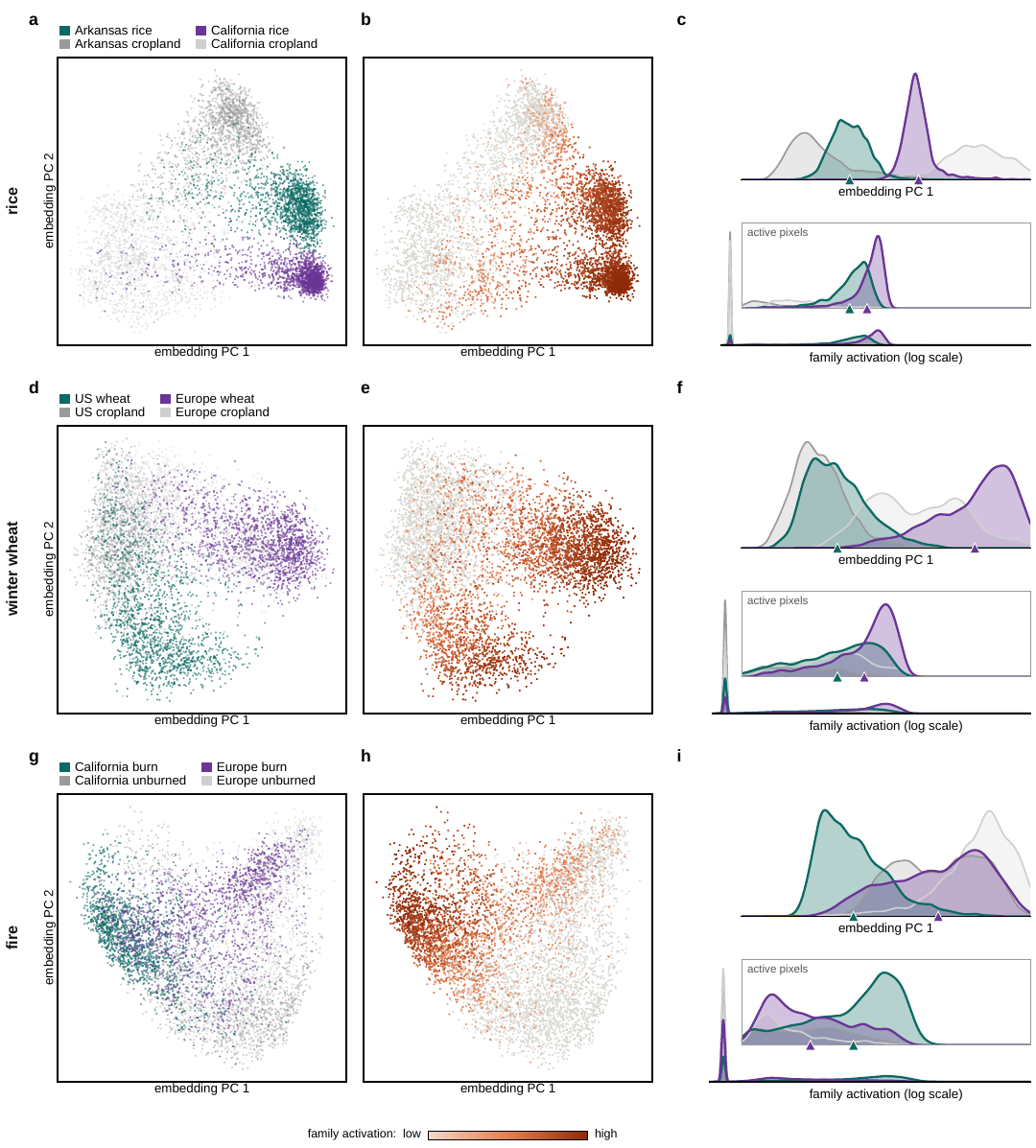}
\caption{\textbf{A feature family found in one region responds to the same concept in another.} Each row takes one feature family in a region, and applies in a new region: rice (Arkansas to Sacramento Valley, California), winter wheat (US southern Plains to France, Germany, Poland and Hungary) and fire (California to Europe). Each row has four groups with equal samples, namely the concept in each region (e.g.\ US wheat, European wheat) and its surroundings in each region (other cropland for the crops, unburned forest for fire).
\textbf{a, d, g,} Raw embeddings of the four groups on their first two principal components.
\textbf{b, e, h,} The same points colored by the family's summed activation (log scale, gray is inactive).
\textbf{c, f, i,} The same four groups along a single line, once in raw embedding space (top) and once in activation space (bottom). Top: position on the first principal component of the group centroids (the direction in which the groups of a differ most). Bottom: the log of the family's summed activation. The inset repeats the family's active curves, magnified vertically, over the same x range. Triangles mark the two concept centroids.}
\label{fig:invariance}
\end{figure}

\textbf{Invariance.} The specificity test asks whether the family does not activate on other disturbances. The invariance test asks whether it still responds to fire, and other families to their concepts, when the embeddings look very different. We tested three concepts, each with a family screened in one region using a dictionary trained only there, and applied without changes to a region the dictionary had never seen (Fig.~\ref{fig:invariance}). For fire, we applied the western US family to European burned pixels. For rice, we applied a family from Arkansas to the Sacramento Valley, and for winter wheat, a family from the US southern Plains to France, Germany, Poland and Hungary. Both crops are grown in similar ways and on similar calendars in the two regions, so a family that transfers is responding to the crop rather than to its surroundings. Crop families were screened on 2018--2021 and evaluated on 2022--2024 in the US, and on the 2018 and 2022 JRC crop-map years in Europe.

Over all 128 dimensions, European wheat is located nearer to European cropland than US wheat in raw embedding space (normalized centroid distance 0.69 against 0.97), and European burned pixels nearer intact forest than Californian burns (0.34 against 0.58). Location explains 39\% of the raw variance among active wheat pixels and 55--58\% among active burned pixels, against 26\% for rice (Table~\ref{tab:invariance}). The families nevertheless respond to the concept in the new region (Fig.~\ref{fig:invariance}b,e,h). Along the family's axis the two regions' concept pixels nearly coincide for rice and wheat, though they lie 2.7 and 3.8 standard deviations apart along the most separating embedding direction (Fig.~\ref{fig:invariance}c,f). Rice and wheat transfer without loss (AUC 0.95 to 0.96 and 0.80 to 0.80).

\begin{table}[h]
\centering
\caption{\textbf{Transfer of three feature families to a new region.} AUC: family activation, concept against its surroundings (other cropland, intact forest). Location $R^2$ is share of raw embedding variance among pixels where the family is active that is explained by location.}
\label{tab:invariance}
\small
\begin{tabular}{llrrr}
\toprule
 & Learned in $\rightarrow$ tested in & AUC, learned & AUC, tested & Location $R^2$ \\
\midrule
Rice & Arkansas $\rightarrow$ Sacramento Valley & 0.95 & 0.96 & 0.26 \\
Winter wheat & US Plains $\rightarrow$ Europe & 0.80 & 0.80 & 0.39 \\
Fire & California $\rightarrow$ Europe & 0.85 & 0.69 & 0.58 \\
\bottomrule
\end{tabular}
\end{table}

\section{Feature ablation}
\label{note:causal}

\textbf{Probes.} We use a standard scikit-learn logistic regression ($C=1$, \texttt{max\_iter} 3{,}000, no class weighting) fitted on frozen embeddings. Pixels are split into train and evaluation halves by a spatial checkerboard of 64-pixel blocks before any fitting, and up to 15{,}000 pixels per class are sampled in each half with a fixed random seed, so classes are balanced. Reported accuracies and AUCs are on the held-out half. The three probe targets are:

\begin{itemize}
    \item Burn history (burned vs unburned pixels within the screening tile)
    \item Terrain aspect (binarized north/south-facing, from Copernicus DEM GLO-30)
    \item Tree vs grass cover (from ESA WorldCover)
\end{itemize}

The latter two are used to test specificity because both are co-resident in the same embedding and are not functions of fire history.

\textbf{Ablation.} A probe is fitted on the embeddings, then re-evaluated on embeddings from which a feature family's contribution has been subtracted as in $x' = x - \sum_{j \in F} f_j(x)\, d_j$, with $d_j$ the decoder direction of feature $j$. The rest of the embedding is left untouched. The control is 200 random feature sets of the same size drawn from the dictionary's active features (activation frequency above 0.5\%), ablated identically. We report their mean and 95\% interval. Both ablations are applied to all three probes, so the burn family must affect the burn probe and not the others, and the random set must affect none of them.

\textbf{Precursor screen.} The disturbance screens require pre-event quiet, which discards any feature that could carry an early-warning signal, so the precursor test inverts that condition, and features are screened for elevation before mapped EFDA bark-beetle and windthrow disturbances (2021--2022 tile-years, where the record reaches back four years) and scored on held-out tile-years. Because EFDA dates a disturbance by the first clear post-event composite, a signal at $Y{-}1$ may be the event itself, and only $Y{-}2$ and deeper count as precursor evidence. Controls are drawn inside the same tile and matched on distance to the nearest prior disturbance, so beetle contagion cannot masquerade as a precursor, and the same family is scored ahead of harvest, which is scheduled and therefore has no biological precursor, as a specificity control.

\end{document}